\documentclass{article}

\usepackage{PRIMEarxiv}
\usepackage{floatrow}
\usepackage[utf8]{inputenc} 
\usepackage[T1]{fontenc}    
\usepackage{hyperref}       
\usepackage{url}            
\usepackage{booktabs}       
\usepackage{amsfonts}       
\usepackage{nicefrac}       
\usepackage{microtype}      
\usepackage{lipsum}
\usepackage{fancyhdr}       
\usepackage{graphicx}       
\graphicspath{{media/}}     
\usepackage{algorithm}
\usepackage{graphicx}
\usepackage{algpseudocode}
\usepackage{multirow}
\usepackage[normalem]{ulem}
\useunder{\uline}{\ul}{}
\newfloatcommand{capbtabbox}{table}[][\FBwidth]
\title{Synthetic data from diffusion models improves drug discovery prediction
}

\author{
  Bing Hu \\
  Computer Science \\
  University of Waterloo \\
  \texttt{bingxu.hu@uwaterloo.ca} \\
  \And
  Ashish Saragadam \\
  School of Public Health Sciences \\
  University of Waterloo \\
  \texttt{asaragadam@uwaterloo.ca} \\
    \And
  Anita Layton \\
  Applied Mathematics \\
  University of Waterloo \\
  \texttt{anita.layton@uwaterloo.ca} \\
   \And
  Helen Chen \\
  School of Public Health Sciences \\
  University of Waterloo \\
  \texttt{helen.chen@uwaterloo.ca} \\
}

\begin{document}
\maketitle

\begin{abstract}
Artificial intelligence (AI) is increasingly used in every stage of drug development.
Continuing breakthroughs in AI-based methods for drug discovery require the creation, improvement, and refinement of drug discovery data. 
We posit a new data challenge that slows the advancement of drug discovery AI: datasets are often collected independently from each other, often with little overlap, creating data sparsity.
Data sparsity makes data curation difficult for researchers looking to answer key research questions requiring values posed across multiple datasets. We propose a novel diffusion GNN model Syngand capable of generating ligand and pharmacokinetic data end-to-end. 
We show and provide a methodology for sampling pharmacokinetic data for existing ligands using our Syngand model. 
We show the initial promising results on the efficacy of the Syngand-generated synthetic target property data on downstream regression tasks with AqSolDB, LD50, and hERG central. 
Using our proposed model and methodology, researchers can easily generate synthetic ligand data to help them explore research questions that require data spanning multiple datasets.
\end{abstract}


\section{Introduction}

There is a growing trend towards leveraging artificial intelligence (AI) in every stage of drug development \cite{kim2021comprehensive}. 
Drug development is an expensive process: it costs \$2-3 billion dollars and 13-15 years to bring a single drug to market. 
Drug discovery AI, by enabling the high-throughput screening (HTS) of ligand candidates, is geared to reduce the developmental costs of drugs by revolutionizing how ligands are designed and tested \cite{pushpakom2019drug}. 
Drug development AI has found great initial success such as in poly-pharmacy \cite{vzitnik2015gene}, drug re-purposing \cite{thafar2022affinity2vec, Morselli_Gysi_2021}, drug-target interaction \cite{lian2021drug}, drug response prediction \cite{pouryahya2022pan}, and in search of new antibiotics \cite{stokes2020deep}. 
Equally important to advances in AI for drug discovery are the equal improvements in available public data for training and testing these models \cite{huang2021therapeutics, guacamol, gaulton2017chembl}.

Only through equal strides in the development and refinement of drug discovery data, and the application of advanced AI models to that data, do breakthroughs happen for AI-based methods for drug discovery. 
Huang et al. \cite{huang2021therapeutics} noted 3 key challenges for drug discovery data to attracting ML scientists to therapeutics: (1) a lack of AI-ready datasets and standardized knowledge representations; (2) datasets scattered around the bio-repositories without curation; (3) a lack of data focused for rare diseases and novel drugs in development. We posit another data challenge that slows the advancement of drug discovery AI: datasets are often collected independently, often with little overlap, creating data sparsity. Data sparsity poses difficulties for researchers looking to answer research questions requiring data values posed across multiple different datasets.  

Although there are advances towards large-scale data repositories containing vast broad data about each ligand \cite{gaulton2017chembl, kim2023pubchem} and cells \cite{jin2020metastasis, huang2021therapeutics}, these large-scale data repositories face unique challenges for scalability and data sparsity. 
Auxiliary data often needs to be gathered for each ligand, such as on pharmacokinetic values of water solubility \cite{sorkun2019aqsoldb}, toxicity \cite{ld50}, and hERG \cite{du2011hergcentral}. 
However, as data repositories become larger, that task becomes increasingly expensive and time-consuming. 
In this paper, we propose Syngand, an end-to-end generative AI diffusion model, that can solve both problems of data sparsity and scalability faced by data repositories. 
Our model is trained to generate novel ligands with synthetic pharmacokinetic data that describe them or fill in pharmacokinetic data when given an existing ligand. 

Recent advances combining Denoising diffusion probabilistic models (DDPMs) \cite{ddpm} with graph-neural networks (GNNs) have resulted in a new class of diffusion models capable of generating ligand structures \cite{guo2023diffusion, digress, diffbridge, difflinker}. Although some use ligand properties to aid in the conditional generation of the ligand \cite{cdgs}, none aim to generate pharmacokinetic data alongside the ligand diffusion pipeline. 
A recent study in diffusion-generated synthetic data \cite{azizi2023synthetic} has shown improvements in performance for ImageNet classification with synthetic augmented real image data.
In the present study, we show that diffusion-generated synthetic ligand data can be used to improve performance on downstream drug discovery regression tasks for water solubility \cite{sorkun2019aqsoldb}, acute toxicity (LD50) \cite{ld50}, and hERG Central \cite{du2011hergcentral}.

Our contributions are as follows:
\begin{itemize}
    \item We propose a novel diffusion GNN model Syngand capable of generating ligand and pharmacokinetic data end-to-end.
    \item We develop a methodology for sampling pharmacokinetic data for existing ligands using our model.
    \item We show promising initial results for the efficacy of the generated synthetic data in augmenting real data for downstream pharmacokinetic tasks for AqSolDB \cite{sorkun2019aqsoldb}, LD50 Toxicity \cite{ld50}, and hERG Central \cite{du2011hergcentral}.
\end{itemize}

\section{Background}

Diffusion methods model complex datasets using families of probability distribution while maintaining computational tractability for learning, sampling, inference and evaluation \cite{guo2023diffusion}. 
The denoising diffusion probabilistic model (DDPMs) approach \cite{ddpm} systematically destroys the structure in the data through a forward diffusion process, and then in a reverse diffusion process, learns how to restore the structure in the data from noise. 
DDPMs have been increasingly used for drug discovery to allow for the rapid generation and high throughput screening (HTS) of drug candidates \cite{gomez2018automatic, de2018molgan}. 

\textbf{C}onditional \textbf{D}iffusion models based on discrete \textbf{G}raph \textbf{S}tructures (CDGS) can be used to generate small-molecule molecular graphs with similar distributions to real small-molecules \cite{cdgs}. 
The CDGS method utilizes a graph noise prediction model built using hybrid message passing blocks (HMPB) \cite{cdgs}, which allows for aggregation of local and global node and edge features. CDGS differs from \cite{jo2022scorebased} which uses separate networks for node and edge denoising. 
E(n)-equivariant GNN (EGNN) \cite{egnn} introduces a new model capable of learning graph equivariant to rotations, translations, and permutations. 
The EGNN architecture is the basis of E(3)-equivariant diffusion model (EDM) \cite{edm} that performs a diffusion process in euclidean space to generate small-molecules. 
An alternative approach based on EGNN is DiffBridge \cite{diffbridge} which uses Lyapunov functions to create prior bridges, injecting physical and statistics information, to guide the diffusion process. Compared to EGNN \cite{egnn} and EDM \cite{edm}, DiffBridge showed improved performance in terms of physical energy and molecular stability during molecule generation \cite{diffbridge}.
Digress \cite{digress} combines discrete diffusion \cite{austin2021structured} and graph transformers \cite{dwivedi2020generalization} for small-molecule generation. 
Digress utilizes a noise model \textit{close to the true data distribution} \cite{digress} that preserves the marginal distribution of nodes and edge types rather than using uniform noise.

In conditional diffusion models \cite{saharia2022photorealistic, ramesh2022hierarchical}, an additional input is available during training and sampling which allows the model to generate data given the conditioning signal. 
Conditional guidance is utilized in Digress \cite{digress} on graph-level properties, such as cycles, and spectral and molecular features, to augment the input with auxiliary conditioning signals to improve training and sampling performance. 
Conditional guidance with target properties such as water solubility, binding affinity, toxicity, drug-likeness, and molecular weight is possible with latent state optimization or Monte-Carlo sampling to optimize molecule generation \cite{richards2022conditional, aumentado2018latent, lee2023exploring, wang2022relation}. Other works use reinforcement learning to generate molecules with specified properties such as weight or solubility \cite{mahmood2021masked, kwon}. The challenge with these approaches is the generalizability of available datasets for these properties to the small-molecule search space. Datasets for important target properties such water solubility \cite{sorkun2019aqsoldb}, LD50 toxicity \cite{ld50}, and hERG central \cite{du2011hergcentral} range between 7k-300k examples, which all pale in comparison to the combinatorial search space of all small-molecules.

Therapeutics Data Commons \cite{huang2021therapeutics}, PubChem \cite{kim2023pubchem}, ZINC \cite{irwin2005zinc}, ChEMBL and associated curations \cite{gaulton2017chembl, guacamol} all aim to provide large data repositories of high-quality ligand data curated for drug discovery. 
Given the costly nature of data collection for drugs, extending additional pharmacokinetic properties across all ligands in a large data repository is expensive.
Data sparsity is actuated by the expensive nature of data collection as only small sets of ligands can be feasibly tested for target property data collection studies. 
Thus, the data sparsity challenge gives rise to barriers for researchers interested in answering key research questions requiring data across multiple datasets. 

Determining the relationship between water solubility, toxicity, and ligand structure is an important research question for drug design \cite{KAWABATA20111, biomedicines10092055}. There are little over 300 ligands that can be collected from public datasets AQSolDB \cite{sorkun2019aqsoldb}, LD50 toxicity \cite{ld50}, and hERG central \cite{du2011hergcentral} for researchers to look to analyze these complex relationships. As 300 ligands may not sufficiently generalize, additional data collection would be necessary. 

In the present study, we aim to solve the data sparsity problem in drug discovery by training a model capable of generating synthetic ligand data.
Our model Syngand extends existing ligand diffusion models by integrating target property generation with ligand generation in a fully end-to-end diffusion pipeline. 
As researchers may want to generate target properties for a fixed set of ligands, we provide a diffusion sampling method that can be used to generate target pharmacokinetic data when given an existing ligand.
Finally, we show in initial experiments the efficacy of the generated synthetic data in augmenting real data for downstream drug-discovery regression tasks on AQSolDB \cite{sorkun2019aqsoldb}, LD50 \cite{ld50}, and hERG central \cite{du2011hergcentral}. 

\section{Methodology}

The model is trained on 1.3 million ligands collected from Guacamol \cite{guacamol}, and training property datasets AqSolDB \cite{sorkun2019aqsoldb}, LD50 toxicity \cite{ld50}, and hERG Central \cite{du2011hergcentral}. Training property datasets are all collected from Therapeutics Data Commons (TDC) \cite{huang2021therapeutics}. We selected a subset of 60k ligands from AqSolDB, LD50, and hERG Central to generate synthetic target properties for existing ligands. 

\subsection{Data Processing}

Guacamol contains around 1.27 million ligands curated from ChEMBL after charge neutralization, and removing salts, molecules with SMILES longer than 100 characters, molecules containing certain elements, and certain molecules for benchmark testing.

The target properties that we will be focusing on are AqSolDB \cite{sorkun2019aqsoldb}, LD50 \cite{ld50}, and hERG Central \cite{du2011hergcentral}.  
AqSolDB \cite{sorkun2019aqsoldb} measures the aqueous solubility of around 9.9k ligands. Poor water solubility can lead to inadequate bioavailability and can be toxic. LD50 \cite{ld50} measures the acute toxicity of around 7.3k ligands. hERG Central \cite{du2011hergcentral} measures the blocking of the human ether-a-go-go related gene (hERG) for around 306k ligands. hERG is crucial for coordinating the beating heart and we use values of hERG inhibition at 1$\mu$M of the ligand. 
We will look to generate synthetic values for these 3 target properties through an end-to-end diffusion and sampling process.
To train our Syngand diffusion network, we first have to construct a training dataset combining data from Guacamol \cite{guacamol} and our 3 target property datasets AqSolDB \cite{sorkun2019aqsoldb}, LD50 \cite{ld50}, and hERG Central \cite{du2011hergcentral}.

In merging the Guacamol dataset \cite{guacamol} with our target property datasets, we first identify common molecules between Guacamol and each target property dataset. As a percentage of each target property dataset, 32.4\% of molecules in AqSolDB exist in Guacamol, 38.7\% of molecules in LD50 exist in Guacamol, and 59.1\% of hERG Central exist in Guacamol.  
After identifying common molecules between Guacamol and each target property dataset, we merge Guacamol and target property datasets AqSolDB, LD50, and hERG on these common molecules. 
Analyzing the merged dataset, of the 1.27 million ligands in Guacamol, 86.6\% of these molecules do not have any target property values. 

To improve the amount and quality of our training data, we look to add molecules in our target property datasets AqSolDB, LD50, and hERG Central that did not previously exist in Guacamol into our training data. As Guacamol contains molecules after careful curation from ChEMBL \cite{guacamol, gaulton2017chembl}, we follow similar filtering rules when adding molecules from target property datasets into our training data. We filter out any molecules from target property datasets that have molecule SMILES strings longer than 100 and molecules with elements besides the 13 included in Guacamol \cite{guacamol}. There are 1.33 million molecules in the training dataset after adding the filtered list of target property molecules, of which 82.7\% of these molecules do not have any target properties. 
\subsection{Syngand Model}

We extend upon the base DiGress model \cite{digress} to create the novel Syngand model that is capable of end-to-end training and diffusion of target property values with ligand diffusion. We accomplish this by introducing an additional continuous diffusion process alongside the DiGress discrete diffusion process for the target properties. The same graph transformer network is used for Syngand as defined in DiGress. In this section, we focus on the formulation and algorithms for target property diffusion as described below.

\subsubsection{Diffusion Process for Target Properties}

Consider target properties $\mathbf{y}$. The diffusion process adds Gaussian noise independently to each target property:
\begin{equation}
q(\bold{y}^t | \bold{y}^{t-1}) = \mathcal{N}(\alpha^{t|t-1}\bold{y}^{t-1}, (\sigma^{t|t-1})^2\bold{I}) 
\end{equation}
This process can be written as:
\begin{equation}
q(\bold{y}^t | \bold{y}) = \mathcal{N}(\bold{y}^t | \alpha^{t}\bold{y}, \sigma^{t}\bold{I}) 
\end{equation}
where $\alpha^{t|t-1} = \alpha^t/\alpha^{t-1}$ and $(\sigma^{t|t-1})^2 = (\sigma^t)^2 - (\alpha^{t|t-1})^2(\sigma^{t-1})^2$.
To obtain a variance-preserving process, the variance chosen is $(\sigma^t)^2=1-(\alpha^t)^2$ \cite{kingma2023variational}. The true denoising process can be computed in closed form:

\begin{equation}
    q(\bold{y}^{t-1}|\bold{y}, \bold{y}^t) = \mathcal{N}(\mu^{t\gets t-1}(\bold{y}, \bold{y}^t), (\sigma^{t \gets t-1})^2 \bold{I})
\end{equation}

with
\begin{equation}
    \mu^{t \gets t-1}(\bold{y}, \bold{y}^t) = \frac{\alpha_{t|t-1}(\sigma^{t-1})^2}{\sigma_t^2}\bold{y}^t + \frac{\alpha^{t-1}(\sigma^{t|t-1})^2}{(\sigma^t)^2}\bold{y} \quad and \quad \sigma^{t\gets t-1} = \frac{\sigma_{t|t-1}\sigma_{t-1}}{\sigma_{t}}
\end{equation}

The denoising network is trained to predict the noise components $\hat{\epsilon}_\bold{y}$ instead of $\hat{\bold{y}}$ itself \cite{ddpm}. The equation is as follows:
\begin{equation}
    \alpha^t\hat{\bold{y}} = \bold{y}^t - \sigma^t\hat{\epsilon}_\bold{y}
\end{equation}

The network is optimized by minimizing the mean squared error between the predicted noise and true noise. Sampling is done similarly to standard Gaussian diffusion models. 

\subsection{Generation of Target Properties for Existing Ligands}

\begin{figure}[h]
\begin{center} 
\includegraphics[width=1\linewidth]{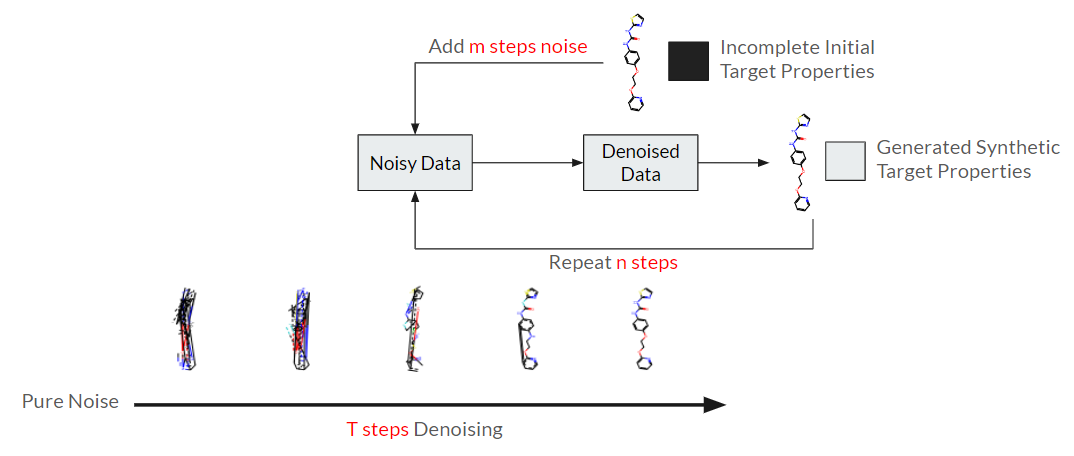} 
\end{center}
\caption{Overview of the methodology for generating target properties for existing ligands. Existing ligards with incompleted initial target properties are repeatedly partially denoised and the generated synthetic target properties are output.}\label{fig:diag}
\end{figure}

We propose a novel generation process to generate target properties for existing ligands (\autoref{fig:diag}). This allows the Syngand model to both be able to generate target property values given an input ligand, as well as to generate new ligands and their target property values. This allows researchers flexibility to directly query target properties from the model for any number of ligands of interest quickly and efficiently. 

\begin{algorithm}
\caption{Generation of target properties for existing ligand}\label{alg:cap}
\begin{algorithmic}
\Require A graph $G = (\bold{X}, \bold{E})$, initial target properties $\bold{y}$, M iterations, N steps where $N<T$
\For {$it = M$ \textbf{to} $1$}
\State Sample $G^N \sim \bold{X}\bar{\bold{Q}}^{N}_X \times E\bar{\bold{Q}}^{N}_E$ \Comment{Sample a discrete noisy graph}
\State Sample $\epsilon_y \sim \mathcal{N}(0, \bold{I}_n)$ \Comment{Sample a Gaussian noise for target properties}
\State $z_y^N \gets \alpha^N(\bold{y}) + \sigma_N(\epsilon_y)$
\For {$t = N$ \textbf{to} $1$}
\State Sample $\epsilon_y \sim \mathcal{N}(0, \bold{I}_n)$
\State $z \gets f(G^t, t)$ \Comment{Structural and spectral features as defined in DiGress}
\State $\hat{p}^X, \hat{p}^E \gets \phi_{\theta}(G^t, z)$ \Comment{Forward Pass}
\State $p_\theta(x^{t-1}_i|G^t) \gets \sum_x q(x^{t-1}_i | x_i = x, x_{i}^{t}) \hat{p}_{i}^\bold{X}(x) \quad i \in 1,...,n$ \Comment{Posterior}
\State $p_\theta(e^{t-1}_{ij}|G^t) \gets \sum_e q(x^{t-1}_{ij} | e_{ij} = e, e_{ij}^{t}) \hat{p}_{ij}^\bold{X}(e) \quad i,j \in 1,...,n$
\State $G^{t-1} \sim \prod_i p_\theta(x^{t-1}_i|G^t)\prod_{i,j}p_\theta(e^{t-1}_{i,j}|G^t)$ \Comment{Categorical distr.}
\State $z^{t-1}_y \gets \frac{1}{\alpha_{t|t-1}}z^t - \frac{\sigma_{t|t-1}^2}{\alpha_{t|t-1}\sigma^t}\phi_\theta(z^t, t) + \sigma_{t \gets t-1}(\epsilon_y)$ \Comment{Reverse iteration for y}
\EndFor
\State $\bold{y} \gets z_y^0$ 
\EndFor
\State \Return $\bold{y}$
\end{algorithmic}
\end{algorithm}

\autoref{alg:cap} shows the pseudocode for the proposed method to generate target properties for existing ligands. This methodology is similar to existing prior bridging techniques as used in DiffBridge \cite{diffbridge}. 
The idea is to repeat a partial denoising process, while bridging in the desired ligand and any initial target property values, until a fixed number of repetitions is reached.
The key parameters are the number of steps $N$ in the denoising process and the number of times $M$ this process is repeated. 
If no existing target properties exist for the input ligand graph $G$, the initial target properties $\bold{y}$ can be imputed to average values. 
Likewise, if the input ligand has pre-existing target properties, the initial target properties $\bold{y}$ can be initialized accordingly.

\section{Results}

\begin{figure}[h]
\begin{center} 
\includegraphics[width=1\linewidth]{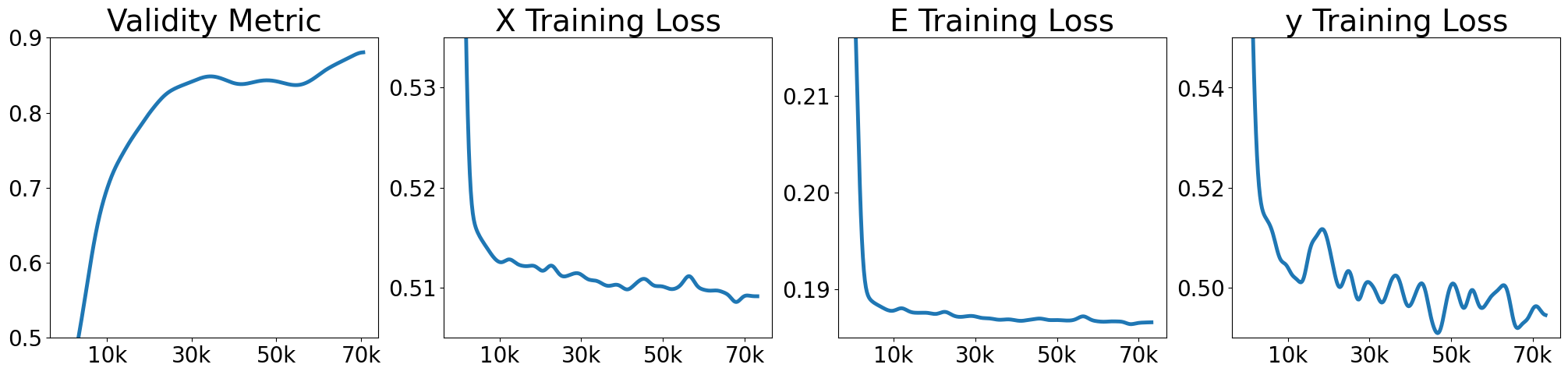} 
\end{center}
\caption{Pre-training loss for ligand graph $G = (X,E)$ and y target properties as well as the relaxed validity metric.}\label{fig:pretraining}
\end{figure}

\autoref{fig:pretraining} shows the pre-training graphs for the Syngand model. The model is trained for 100 epochs and validated after every epoch. The model is trained until convergence with vertex loss around 0.51, edge loss around 0.18, target property loss around 0.49, and validity around 88\%. Relaxed validity is computed using RdKit \cite{landrum2013rdkit} over 64 generated synthetic ligands. The pre-trained model is used for our experiment on generating synthetic target property values for select input ligands using~\autoref{alg:cap}. 

\subsection{Generated Synthetic Target Properties}

A set of 60k ligands are selected from AqSolDB \cite{sorkun2019aqsoldb}, LD50 \cite{ld50}, and hERG Central \cite{du2011hergcentral}. The pre-trained model and algorithm~\autoref{alg:cap} is then applied to the selected ligands to generate synthetic target property values. We utilize parameters of $N=50$ (iterations) and $M=10$ (steps) for~\autoref{alg:cap}.
The generated synthetic dataset, containing 60k ligands with AqSolDB, LD50, and hERG values, can be used to augment real data in research work needing large quantities of ligands with data spanning these three datasets. 

\begin{figure}
\begin{center} 
\includegraphics[width=1\linewidth]{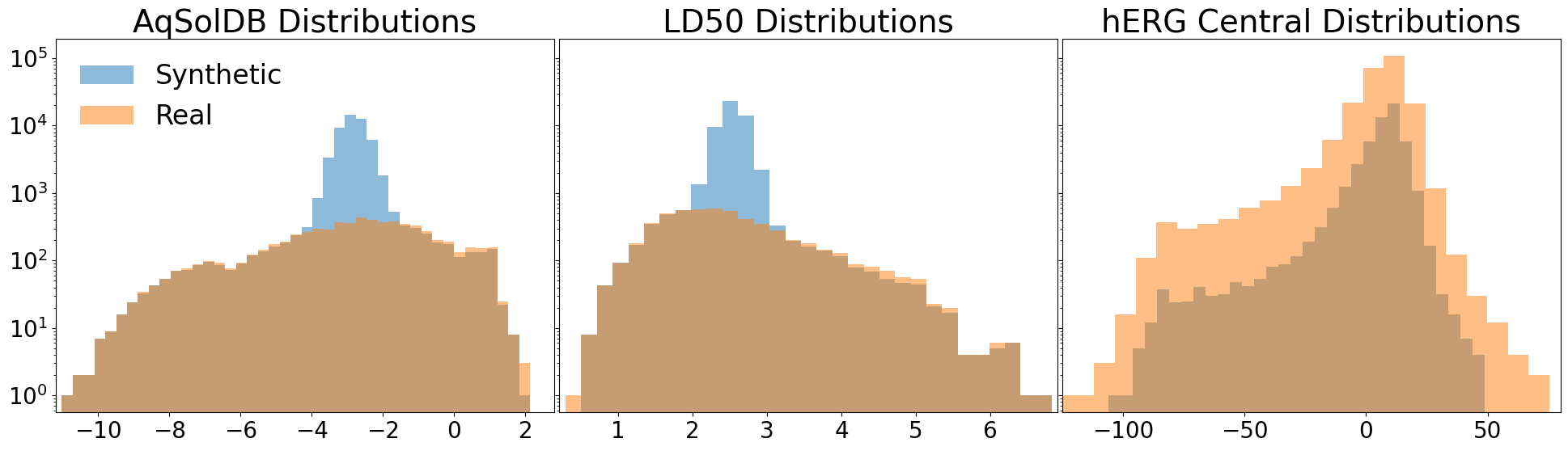} 
\end{center}
\caption{Distribution of generated synthetic target properties, AqSolDB (left), LD50 (middle), and hERG (right) for selected ligands. Synthetic distributions are in blue while real distributions are in yellow. Log scale is used for the y-axis.}\label{fig:dist}
\end{figure}

\autoref{fig:dist} shows the distribution of the generated synthetic target properties for a set of 60k selected input ligands. 
Hellinger distances, 0.43, 0.50, and 0.08, are computed to compare the distribution of synthetically generated target properties with real target properties for AqSolDB, LD50, and hERG respectively. 


\begin{table}
\begin{tabular}{lcc}
\hline
Data    & \multicolumn{1}{l}{Real Mean (STD)} & \multicolumn{1}{l}{Synthetic Mean (STD)} \\ \hline
AqSolDB & -2.91 (2.30)                        & -2.85 (0.87)                             \\
LD50    & 2.53 (0.95)                         & 2.53 (0.30)                              \\
hERG    & 6.23 (11.18)                        & 7.22 (9.55)                              \\ \hline
\end{tabular}
\caption{Comparing mean and variance values between real and synthetic target property values.}\label{tab:mean}
\end{table}
\autoref{tab:mean} compares the mean and variance of the real and synthetically generated target property values. 
The mean of the generated synthetic data aligns closely with real data means with deviations between 2-16\%. Synthetic data has variance much different from real data differing by a factor of 1.17-3.16x.
The mean and variation of generated synthetic data are affected by the noise sampling strategy, in our case Gaussian, and sampling hyperparameters of the number of iterations ($M=50$) and steps ($M=10$). 

A key limitation is that the generated data remains clustered around the mean.
Future work will look to optimize the noise sampling strategy and sampling hyperparameters to improve the univariate quality of the generated synthetic data compared to the real data. Additional pairwise correlations and multivariate statistical analysis will be conducted in future work to further evaluate the quality of the generated synthetic data. 




\subsection{Downstream Drug Discovery Regression Tasks}

After the generation of synthetic target properties for our selected ligands, the quality of the generated target properties is measured by the performance of the synthetic data on a downstream task. 
In this experiment, we aim to measure the machine learning efficiency (MLE) \cite{borisov2022language} of the generated synthetic data.
Machine learning efficiency (MLE) represents the ability of the generated data to replace or augment real data in the training process. MLE evaluates the performance of discriminative models trained on synthetic data. To measure MLE, the models are tested on real test data, and the scores are compared to the original performance when models are trained on real data \cite{borisov2022language}. 

\begin{table}[h]
\begin{tabular}{llccc}
\hline
\multicolumn{1}{c}{}     & \multicolumn{1}{c}{} & \multicolumn{3}{c}{Trainset-Testset}                                                       \\
Data                     & Metric               & \multicolumn{1}{l}{Aug-Aug} & \multicolumn{1}{l}{Aug-Real} & \multicolumn{1}{l}{Real-Real} \\ \hline
\multirow{4}{*}{AqSolDB} & MSE                  & 0.172                       & {\ul \textbf{0.399}}         & 0.927                         \\
                         & MAE                  & 0.329                       & {\ul \textbf{0.632}}         & 0.753                         \\
                         & R2                   & 0.018                       & {\ul \textbf{0.053}}         & -1.199                        \\
                         & PCC                  & 0.149                       & {\ul \textbf{0.263}}         & 0.19                          \\ \hline
\multirow{4}{*}{LD50}    & MSE                  & 0.029                       & {\ul \textbf{0.072}}         & 0.345                         \\
                         & MAE                  & 0.135                       & {\ul \textbf{0.226}}         & 0.459                         \\
                         & R2                   & 0.075                       & {\ul \textbf{0.017}}         & -2.292                        \\
                         & PCC                  & 0.279                       & {\ul \textbf{0.151}}         & 0.094                         \\ \hline
\multirow{4}{*}{hERG}    & MSE                  & 29.8                        & {\ul \textbf{34.0}}          & 34.7                          \\
                         & MAE                  & 4.32                        & {\ul \textbf{4.76}}          & 4.81                          \\
                         & R2                   & 0.01                        & {\ul \textbf{0.002}}        & -0.018                        \\
                         & PCC                  & 0.129                       & {\ul \textbf{0.102}}        & 0.098                         \\ \hline
\end{tabular}
\caption{Comparing drug discovery regression performances between different combinations of augmented and real train sets, and augmented and real test sets. Values are averaged over 30 trials with the best scores on the real test set bolded. The underlined values are statistically significant. Hyphenated pairs correspond to the train and test set used in the experiment.  }\label{tab:res}
\end{table}

For this experiment, we develop a Linear Regression (LR) and XGBoost (XGB) model for AqSolDB \cite{sorkun2019aqsoldb}, LD50 \cite{ld50}, and hERG central \cite{du2011hergcentral} to evaluate our generated synthetic augmented real data for downstream drug discovery tasks.
The model takes as input learned embeddings from ChemBERTa \cite{chithrananda2020chemberta} of ligand SMILES to predict desired target property values using the discriminative models. 


To prevent data leakage, we follow a procedure to first 
divide real and synthetic data before combining them to form train and test sets.
We divide synthetic data into segments denoted $A_{s}$ and $B_{s}$ using a 85\%/15\% split. We divide real data into segments denoted  $A_{r}$ and $B_{r}$ using a 50\%/50\% split.
The real train set is defined as $A_{r}$ and the real test set is defined as $B_{r}$. The augmented train set is defined as $A_{r} \cup A_{s}$ and the augmented test set is defined as $B_{r} \cup B_{s}$.
Outliers are removed from both real and augmented train and test sets based on $\pm1.5$IQR bounds on the synthetic data. 


\autoref{tab:res} shows the results of downstream task performance of using a synthetic augmented real dataset compared to just the real dataset.
With our simple model, we show that an augmented dataset can outperform real data with statistical significance over 30 runs after removing outlier values.
Additional models and downstream tasks will be explored in future work.

\section{Conclusion}

In this paper, we 
developed Syngand, an end-to-end model and methodology that can generate novel ligands and their corresponding target properties. Additionally, we propose a novel method for generating target property values given an existing ligand using our Syngand model. 
Finally, we show the efficacy of the generated synthetic ligand data by comparing performance on downstream drug discovery regression tasks with AqSolDB, LD50, and hERG. 
One key limitations of this work is the poor univariate quality of the synthetic data being too clustered compared to the real data. Further hyperparameter optimizations will need to be explored to improve on the quality of the generated syntehtic data. Other limitations   are to expand the scope of the research to also measure the multivariate quality of the generated synthetic data, and to expand downstream testing of the synthetic data to include additional tasks and models.  
Although there are key limitations that still need to be addressed, our methods show promise in tackling the data sparsity problem in drug discovery. 

Future work will look to improve our methods, scale to additional target properties, generate more synthetic data and create better evaluations of this new class of synthetic data. 
A crucial direction for this research is to improve the quality of the generated synthetic data to extend our results to more complex downstream tasks and models. Additional ablation and configuration studies will be explored in future work.
Drug discovery researchers can immediately benefit from Syngand given how easily and quickly they can generate data they need to investigate previously infeasible research questions that span multiple datasets.
Through addressing data sparsity in drug discovery, our methods work to accelerate drug discovery research and drug discovery AI.

\bibliographystyle{unsrt}  
\bibliography{references}

\end{document}